\documentclass[sigplan,10pt]{acmart}
\renewcommand\footnotetextcopyrightpermission[1]{} 
\pdfoutput=1

\acmConference[arXiv]{1}{1}{1}
\acmYear{2019}
\startPage{1}
\setcopyright{none}
\usepackage{booktabs}   
\usepackage{subcaption} 
\usepackage{enumerate}  
\usepackage{listings}
\newcommand{\codelistsize}{\small}
\newcommand{\specname}{\emph}

\begin{document}
\title[AAD]{An Evolutionary Framework for 
Automatic and Guided Discovery of Algorithms 
}

\author{Ruchira Sasanka}
\affiliation{
  \institution{Intel Corporation}            
}
\email{ruchira.sasanka@intel.com}          

\author{Konstantinos Krommydas}
\affiliation{
  \institution{Intel Corporation}            
}
\email{konstantinos.krommydas@intel.com}          

\begin{abstract}
This paper presents \emph{Automatic Algorithm Discoverer} (AAD), an
evolutionary framework for synthesizing programs of high complexity. 
To guide evolution, prior evolutionary algorithms have depended on fitness (objective) 
functions, which are challenging to design. 
To make evolutionary progress, instead, AAD employs {\em Problem Guided Evolution} (PGE),
which requires introduction of a {\em group} of problems together.
With PGE, solutions discovered for simpler problems are used to solve more 
complex problems in the same group. PGE also enables several
new evolutionary strategies, and naturally yields to High-Performance 
Computing (HPC) techniques.

We find that PGE and related evolutionary strategies enable AAD to 
discover algorithms of similar or higher complexity relative
to the state-of-the-art. Specifically, AAD produces Python code for 29
array/vector problems ranging from min, max, reverse, to more challenging
problems like sorting and matrix-vector multiplication.
Additionally, we find that AAD shows adaptability to 
constrained environments/inputs and demonstrates
 \emph{outside-of-the-box} problem solving abilities.

\end{abstract}



\maketitle

\section{Introduction}
\label{aad:introduction}

Program synthesis involves automatically assembling a program 
from simpler components. 
It is analogous to searching the entire space created by all possible
permutations of those components, looking for solutions that satisfy 
given requirements. Many such search strategies (such as enumerative,
deduction-based, constraint-solving, stochastic) have been proposed to
address this
challenge~\cite{flashMeta,SolarThesis,Srivastava2013,MLPBE,ML2,deepCoder,neural1}.

In this work, we propose an evolution-based search strategy, implemented
in the \emph{Automatic Algorithm Discoverer} (\emph{AAD}). AAD can
synthesize programs of relatively high complexity (including loops,
nested blocks, nested function calls, etc.), based on a subset of Python
as grammar, and can generate executable Python code. In this paper we
use AAD to discover algorithmic solutions to array/vector problems. 

Evolutionary algorithms use a {\em fitness} (objective) function to pick 
the fittest individuals from a population~\cite{geneticBook,Koza1994,genetic2,genetic3}.
The traits of the fittest individuals can recombine (cross-over) to form the next generation.
However, designing an effective
fitness function could be challenging for complex problems~\cite{fitness11,fitness12,progSynthNow}.
We propose an alternative way to guide evolution without 
a fitness function, by adding several 
{\em potentially related problems} together into a {\bf group}.
We call this {\em Problem Guided Evolution} {\bf (PGE)} and it is analogous 
to the way we teach children to solve complex problems. For instance,
to help discover an algorithm for finding the area of a {\em polygon},
we may ask a student to find a way to calculate the area of a {\em triangle}.
That is, simpler problems guide solutions to more complex ones.
Notably, PGE does not require knowing the exact algorithm nor the exact
constituents to a solution, but rather a few {\em potentially} related {\em problems}. 
In AAD, PGE allows more complex solutions to be derived through
(i) {\em composition}  (combining simpler ones), and through 
(ii) {\em mutation} of alredy discovered ones.  

Grouping related problems for PGE, like designing a fitness function, 
is not automatic and currently requires human insight. 
However, PGE could be a more natural way to attack complex problems.
As a concrete example, Figure~\ref{fig:sortAlgo} shows code that AAD
produced in order to sort a non-empty array in ascending order
(\texttt{SortAsc}).
To solve this, we grouped ten {\em potentially} related problems 
together to guide evolution: min, max, first/last index, reverse, 
remove first/last, is-in-array, and sort ascending/descending.
AAD used solutions it found itself for {\em some} of those problems
to discover an algorithm for sorting: 
by first finding the minimum of the input array, appending that
minimum to a new list, removing the minimum from the input array, and
repeating these steps until the entire input array is processed. Though
not the most elegant nor the most efficient implementation, a machine
being able to discover an algorithm for sorting starting from a basic
subset of Python illustrates the capabilities of AAD and the utility of PGE. 

Overall, this paper makes the following contributions:

\begin{itemize}

\item Use of {\em Problem Guided Evolution} to eliminate objective functions
      in evolutionary algorithms.

\item Use of multiple evolutionary strategies (such as
      \emph{diverse environments \& solutions},
      \emph{cross-pollination}, and \emph{ganged evolution}), 
      and evaluation of their effectiveness via a wide range of
      experiments.

\item Application of AAD to solve 29 array/vector problems in
      general-purpose Python language, demonstrating evolutionary
      algorithms are capable of solving complex state-of-the-art problems.

\item Support of {\em loops} to discover 
      algorithms that can accept {\em any} (non-zero) {\em input size}.

\item Mapping of inherently parallel evolutionary process to HPC
      hardware and techniques. 

\end{itemize}

\codelistsize
\begin{figure}
\begin{verbatim}
def Min(arg0):
        arr_10 = arg0.copy()
        num_11 = arr_10.pop(0)
        for num_12 in tuple(arr_10):
                bool_14 = (num_11 > num_12)
                if (bool_14):
                        num_11 = num_12
                num_11 = num_11
        return (num_11)

def LastIndOf(arg0, arg1):
        arr_10 = arg0.copy()
        num_11 = arg1
        for num_12, num_13 in enumerate(tuple(arr_10)):
                bool_15 = (num_13 == num_11)
                if (bool_15):
                        arr_10.append(num_12)
                num_13 = arr_10[-1]
        return (num_13)

def RemoveL(arg0, arg1):
        arr_10 = arg0.copy()
        num_11 = arg1
        num_16 = LastIndOf(arr_10, num_11)
        num_14 = arr_10.pop(num_16)
        bool_12 = (num_16 < num_14)
        return (arr_10)

def SortAsc(arg0):
        arr_10 = arg0.copy()
        arr_16 = list()
        arr_14 = arr_16.copy()
        for num_12 in tuple(arr_10):
                num_15 = Min(arr_10)
                arr_14.append(num_15)
                arr_10 = RemoveL(arr_10, num_15)
        return (arr_14)
\end{verbatim}
\caption{AAD synthesized algorithm for sorting}
\label{fig:sortAlgo}
\end{figure}
\normalsize

We find that PGE and related evolutionary strategies are effective at
discovering solutions to our array/vector problems. Among other findings,
notable is the adaptability of AAD to constrained environments and
inputs, as well as its ability to find {\em creative} solutions. 

The rest of the paper is organized as follows: In
Section~\ref{aad:design} we present the design details of AAD.
Specifically, we discuss:
(i) the three constituent parts of AAD, 
with special emphasis on the  \emph{Evolver} and its three phases that
construct the solution, (ii) the \emph{evolutionary strategies} employed
by AAD and similarities to biological evolution, and (iii) engineering
challenges we faced and their solutions, as well as our HPC-oriented
approach in the AAD implementation.
Section~\ref{aad:methodology} presents our experimental setup and
Section~\ref{aad:results} discusses the results of putting AAD in
test in a broad range of experiments encompassing a variety of
problems. 
Last, we present related work (Section~\ref{aad:relatedwork}), a
discussion and future work (Section~\ref{aad:discussion}), and conclude
the paper in Section~\ref{aad:conclusion}.

\section{Design}
\label{aad:design}

\begin{figure}
  \includegraphics[width=0.4\textwidth]{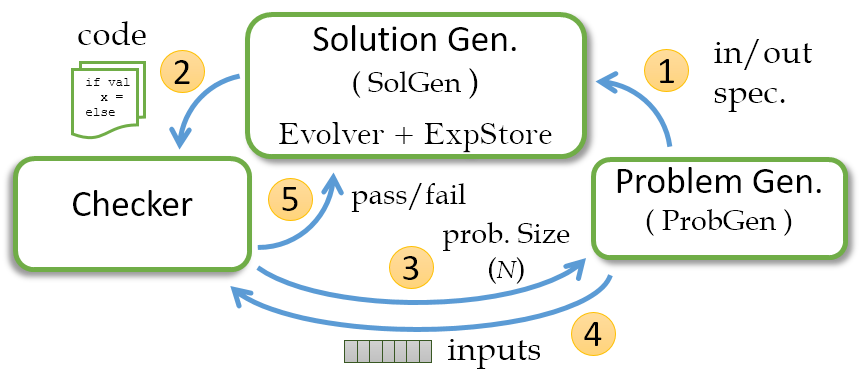}
  \vspace{-0.1in}
  \caption{Components of AAD.}
  \label{fig:components}
\vspace{-0.17in}
\end{figure}

As shown in Figure~\ref{fig:components}, AAD consists of three
components: (i) a Problem Generator (ProbGen) to generate a problem,
(ii) a Solution Generator (SolGen) to come up with potential solutions
(programs), and (iii) a Checker to check those solutions. 

\subsection{Problem Generator (ProbGen)}
\label{sec:probgen}

Each problem we want solved starts with its own ProbGen.  ProbGen is
responsible for: (i) specifying the number and types of inputs and
outputs, and (ii) generating inputs for a given problem size. For
instance, for maximum finding (\texttt{Max}), ProbGen specifies that \texttt{Max} takes
one array as its input and produces a number as its output. In addition,
when requested to generate inputs for a problem of size $N$, it produces
an input array filled with $N$ numbers.

\subsection{Checker}
\label{sec:checker}

Checker is responsible for accepting or rejecting a solution generated
for a given problem. Checker executes the generated program with
input(s) generated by ProbGen, and produces output. The Checker contains
logic to either accept or reject the output, as in~\cite{bitvector1}.
Therefore, a Checker is specific to a given ProbGen, and both go hand in hand.

A Checker does not necessarily need an implementation of the algorithm
it seeks to discover, although some problems do require an
alternative implementation. For instance, the Checker for problem
``Sort'' does not have to sort the input array. Rather, it can compare
each two adjacent elements in the output array and see whether the two
elements are in the expected order. As soon as it detects an out of
order pair, it can declare failure. If each pair of elements is in
order, and the output array contains exactly the same elements as the
input array, which can be checked by removing matching elements, the
Checker can accept the solution.

For some problems based on the physical world, the input and output data
may be found out empirically. For instance, to develop an algorithm that
can predict future temperatures at a specific place, historical
temperature data may be used or data may be gathered using sensors. In
other words, the physical world can function as a Checker for some of
the models (algorithms) we want to discover.

\subsection{The Solution Generator (SolGen)}
\label{sec:solgen}

SolGen primarily consists of two components: 
(i) an Expression/Idiom Store, and (ii) an Evolver.

\subsubsection{Expression/Idiom Store (ExpStore)}

SolGen constructs source programs using a grammar, as
in~\cite{flashFill,pldi18,rosette,sygusPaper}. The subset of Python
grammar AAD uses is stored in ExpStore, and is given in
Table~\ref{tab:grammar}.

In AAD, grammar rules are augmented with type information, as
in~\cite{types1,types2,types3}. AAD supports four types: numbers (NUM), Boolean
(BOOL), arrays (ARR), and arrays-of-arrays (AoA), which can model
matrices. Further, each operand of an expression is marked as a Consumer
(read-only), a Producer (write-only), or a ProdCon (read-modify-write).
With this additional information, a typical binary addition rule becomes:

\centerline{\textsf{NUM (Prod) = NUM (Cons) + NUM (Cons)}}
\vspace{0.05in}

In Table~\ref{tab:grammar}, the producer operands are italicized and
ProdCon operands are underlined. The rest is all consumers. Some
auxiliary grammar rules (e.g., for concatenating statements, function declarations)
are omitted for brevity.

When AAD uses a statement with a BLOCK (a code block), it inserts at
least one randomly selected expression into the BLOCK as a heuristic. We call such a
construct an {\bf idiom}. In addition, when an \specname{If Stmt} is inserted,
an additional expression producing a BOOL is inserted before it, to make
sure that there is an expression producing a BOOL for the \specname{If
Stmt} to consume. Consequently, every \specname{For Stmt} and
\specname{If Stmt} is inserted as an idiom. Reduction is another such
idiom. Idioms allow for faster construction of useful programs. 

In ExpStore, the operands of expressions are generally not given any
identifier names. However, in idioms, if two expressions have a producer-consumer
relationship (e.g., BOOL expression in \specname{If idiom}), we assign the common
operand a common integer identifier to link the producer and the
consumer.

\small
\begin{table}
\begin{tabular}{l  l }

{\bf Expr/Stmt}    & {\bf Representation} \\

 Arithmetic         & {\em NUM} = NUM op NUM   \\
                     & op = $+, -, *, //$ \\

 Compare            & {\em BOOL} = BOOL op BOOL \\
                     & op = $<$, $>$, $==$, $<=$, $>=$, $!=$   \\

 Head/Tail         & {\em NUM} = ARR{\bf [0]} $|$ ARR{\bf [-1]} \\

 Pop \footnotesize{(Head/Tail)} & {\em NUM} = \underline{ARR}{\bf .pop(0)} \\
                 & {\em NUM} = \underline{ARR}{\bf .pop()} \\

 Pop at Ind        & {\em NUM} = \underline{ARR}{\bf .pop(}NUM{\bf)} \\

 Append             &  \underline{ARR}\bf{.append()}     \\

 New Array          & {\em ARR} = {\bf list()}  \\

 Constant          & {\em NUM} = 0 $|$ 1  \\
 Array copy         & {\em ARR} = ARR{\bf .copy()}   \\

 AoA copy           & {\em AoA} = AoA{\bf .copy()}  \\

 Func Arg           &  {\em ANY\_TYPE} = \emph{param}  \\

 Assign Stmt        & {\em NUM} = NUM \\

 Return Stmt        & {\bf return} ANY\_TYPE \\
 
 Reduction          & \underline{NUM} += NUM \\

 If Stmt            & {\bf if} BOOL{\bf :} BLOCK \\

 For Stmt           & {\bf for} {\em NUM} {\bf in} ARR{\bf :} BLOCK\\

                      & {\bf for} {\em NUM{\bf,}  NUM} {\bf in} {\bf enumerate}(ARR){\bf :} BLOCK\\

                      & {\bf for} {\em ARR} {\bf in} AoA{\bf :} BLOCK\\

                      & {\bf for} {\em NUM{\bf,} ARR} {\bf in} {\bf enumerate}(AoA){\bf :} BLOCK\\

 BLOCK              & BEGIN Statements END  \\
\end{tabular}
\vspace{0.05in}
\caption{Expressions and statements used in AAD. ANY\_TYPE represents 
  any of NUM, ARR, BOOL, or AoA. Token \emph{param}
  represents a function parameter.}
\label{tab:grammar}
\vspace{-0.35in}
\end{table}
\normalsize

\specname{For Stmts} allow us to iterate over two types of data
structures -- ARR or AoA. The latter type is used in a context-sensitive
way -- if and only if the function has a parameter of type AoA. For each
of those two types, AAD allows enumerated and non-enumerated
for-loops. In Python, enumerated iterative loops provide both the index
and the item at that index, making them more versatile. However, such a
statement leads to two producers, one for the element and one for the
index. 

In AAD, the \specname{Expr} class representing expressions are
structured so that it can represent multiple Producers, Consumers and/or
ProdCons. A function is modeled as a \emph{sequence} of \specname{Expr}
objects.

For a BLOCK, we add BEGIN and END delimiters  (dummy \specname{Expr}
objects), which are useful in analysis. They are not emitted as part of
generated Python code, though they decide indentation of emitted code.

Since function calls are expressions, ExpStore can include calls to
library functions. However, to be as minimalistic as possible, we use
library calls needed only for basic array access -- pop head/tail or a
given index, and append to tail.

At first glance, ProdCon operands may appear as an unnecessary
complication. However, Python's library functions like {\em append()} 
modify the source operand, and so does the \specname{Reduction}
expression in Table~\ref{tab:grammar}. ProdCon operands allow modeling
these succinctly. 
In program synthesis such operands provide a less obvious but
important benefit because they reduce the total number of operands in an
expression and reduce the number of unique producers in a program,
leading to a reduction of search space.

More expressions and statements can be added to ExpStore as needed, but
only those we use for current results are listed in
Table~\ref{tab:grammar}. Especially, the grammar shown leads to regular
control-flow and avoids statements like {\em break} and {\em continue},
though addition of these is readily supported.

\subsubsection{Evolver}
\label{sec:evolver}

The Evolver is responsible for combining expressions and idioms to
assemble a program (a function), which can potentially solve the problem
presented by ProbGen. The Evolver constructs the solution function ({\bf
SolFunc}) in three phases.

\bigbreak
\begin{flushleft}  
{\bf Phase 1: Building of SolFunc}
\end{flushleft}  

First, Evolver initializes SolFunc so that the input argument
expressions are at the top of the function and the return statement is
at the bottom of the function as shown in Figure~\ref{tab:linking}. The
building of SolFunc boils down to inserting items in ExpStore between
the input arguments and the return statement. Second, the Evolver builds
SolFunc bottom-up, starting from the return statement. If the return
statement consumes a value of type $T$, the Evolver randomly picks an
expression, $E$, from ExpStore that has a producer (or a ProdCon) of type
$T$ and inserts it above the return statement. Now, $E$, has its own
consumer operands, for which producers must be found. Consequently, the
Evolver picks another expression at random from ExpStore to produce each
source operand of $E$, and inserts it randomly somewhere above $E$, but
below function arguments. Instead of selecting an expression, the
Evolver may randomly choose to insert an idiom from the ExpStore. 

If input arguments are only of one type, $T$, just below the input
arguments, the Evolver inserts an expression consuming an operand of
type $T$ and producing a type other than $T$. For instance, if the sole
input argument is an array, the Evolver inserts a randomly picked
expression that consumes an array and produces a number. This heuristic
ensures that values of both types, ARR and NUM, are generated at the top
of the function, allowing any later expressions to consume values of
these two common types.
Additionally, local copies are made of incoming ARR function parameters
to prevent them from being modified within the function (see
Figure~\ref{tab:linking}). 

\bigbreak
\begin{flushleft}  
{\bf Phase 2: Linking Producers and Consumers}
\end{flushleft}  

In Phase 1, Evolver gives each producer operand a unique integer ID,
when an expression is inserted into SolFunc. In Phase 2, consumer (and
ProdCon) operands have to be assigned one of those IDs, linking a
producer and a consumer. 
The Evolver, starts this process at the bottom of SolFunc, starting with
the return statement. The return statement, $R$, has only one consumer
of type $T$. The Evolver looks for all expressions above $R$, for one
producing an operand of type $T$. Out of all those expressions, the
Evolver picks one at random and assigns the ID of that producer (or
ProdCon) to the consumer, thereby linking the two. 
The Evolver continues this process, from bottom to top of SolFunc, until
every consumer operand is linked with a producer (or ProdCon). 
One producer can be consumed by one or more consumers (or ProdCons).

Since AAD supports block nesting, while linking producers and consumers
Evolver has to make sure scoping rules are met. For instance, in
Figure~\ref{tab:linking}, producer \texttt{num\_13} is in an inner block
than its consumer, the return statement, violating scoping rules. There
are several ways to fix this and we chose to alias \texttt{num\_13} with
another producer (e.g., \texttt{num\_11}) at the same level as the
consumer, causing to emit \texttt{num\_11} instead of \texttt{num\_13},
for all operands in all expressions.
If there are multiple consumers of a producer, the producer must be
at the same level as the outermost consumer.

The linking phase also opportunistically detects dead expressions, which
are those with producers that are not consumed. A rigorous attempt is
{\em not} made to detect and remove all dead code because Python
interpreter executing the produced code can do this for itself.

At the end of Phase 2, SolFunc is complete and can be compiled and
executed.

\codelistsize
\begin{figure}
\begin{verbatim}
def Func(arg0):
   arr_10 = arg0.copy()
   num_11 = arr_10[-1]
   ...
   for num_12 in tuple(arr_10):
       ...
       if (bool_14):
             num_13 = ...
       ...       
    return (num_13)
\end{verbatim}
\caption{Linking producers and consumers}
\label{tab:linking}
\end{figure}
\normalsize

\bigbreak
\begin{flushleft}  
{\bf Phase 3: Operator \& Function Call Mutation }
\end{flushleft}  

The completed SolFunc can be optionally mutated in Phase 3. The first
four expressions in Table~\ref{tab:grammar} are designed to capture
multiple operations. For instance, the first grammar rule for binary
\specname{Arithmetic} operations captures four operations: +, -, *, //
(integer division).

As a result, we can easily change an expression from an addition to a
multiplication without re-building SolFunc or re-linking producers and
consumers. In Phase 3, the Evolver randomly changes these operations. In
addition to operators, Phase 3 can mutate an existing function call 
(e.g., \texttt{Max}) to another compatible call (e.g., \texttt{Min}), with the same type
of argument(s) and the return type.

\subsection{Checking Output}
\label{design:checking}

Once SolFunc is built (and mutated), it is executed to produce output 
using Python's \texttt{exec()} function. The output is checked with the
Checker, which either accepts or rejects the output. If the first output
is accepted, the same SolFunc is tested with more inputs of different
sizes, generated using ProbGen. If all those tests are accepted by the
Checker, the SolFunc is declared a solution for the problem. The above 
three phases constitute one evolutionary {\bf step}.

\subsection{Evolutionary Strategies}
\label{design:strategies}

This section describes evolutionary strategies used by AAD.

\subsubsection{Composition}

AAD uses
self-discovered solutions to simpler problems to 
compose more complex solutions. 
To this end, AAD evolves an entire {\bf group} of problems at once, as shown in
Figure~\ref{fig:ranks}(a). Once an acceptable SolFunc is found for one
problem in the group, it is allowed to be called by others by adding an
appropriate function call to ExpStore. Since a function call is an
expression, when a SolFunc is accepted by the Checker, AAD creates a
function call expression for it with appropriate parameters. AAD uses
the input-output description given by ProbGen to determine the type of
each parameter and the return type. By AAD's convention, the input
parameters are always read-only (consumers). However, a function like
\texttt{Remove(\underline{ARR}, NUM)} modifies the first parameter. We
allow such functions to be created as well by allowing ProbGen to
identify the first parameter as a ProdCon and omit a separate return
value (i.e., omit a separate producer). When emitting, AAD emits such a
function with the same identifier for the return value and the first
argument as \texttt{{\em arr1} = Remove({\em arr1}, num1)}.

Function composition has a profound effect on reducing the size of
search space. For instance, assume we allow $N$ statements in SolFunc
and the ExpStore contains $E$ items to pick from. Since each statement
in SolFunc can be filled in $E$ ways (with repetition), there are $E^N$
unique functions we can create. This is the size of the search space. If
a problem requires two functions of size $N$ (one function calling the
other), without function composition we may need up to $2N$ statements
to solve this problem. That increases the search space to $E^{2N}$
possibilities. In contrast, if we have an additional function, we have
$E+1$ expressions in the ExpStore. Therefore, the number of unique
SolFuncs we can create becomes $(E+1)^N$. For non-trivial cases,
$$E^{2N} >> (E+1)^N.$$ 
Therefore, function composition is much more effective at reducing
search space than allowing more statements in SolFunc. 

Although genetic mutation and recombination get the most attention, 
composition can be also seen in biological cell evolution, where it is
called endosymbiosis~\cite[p.~77]{lifeOfBio}. For instance, it is widely
believed that mitochondria present in eukaryotic cells (like animal or
plant cells) have been captured from the primitive environment, where
mitochondria existed independently as a more primitive prokaryotic cell
(like bacteria). However, for mitochondria to evolve as a prokaryot, it
must have solved some natural challenge (problem). In fact, it solved
the problem of energy production on its own and is the `power plant' in
a eukaryotic cell. This shows that having many problems to solve is a
key to evolution. This is the main {\em insight} used by AAD; if we want to
solve larger problems, there should be many other simpler problems
present, to guide evolution. 

\subsubsection{Ganged Evolution}

Related problems, which have the same number and types of inputs and
outputs (e.g., \texttt{Min} and \texttt{Max}, or \texttt{SortAsc} 
and \texttt{SortDesc}), are ganged together
into a single {\bf gang} (see  Figure~\ref{fig:ranks}(a)). Once a
SolFunc is generated for one of the problems in the gang, at the end of
Phase 2 or 3, it is tested on \emph{all} problems in the gang. This allows
solutions to be found faster because a potential solution (SolFunc) may
satisfy one of the many problems in the gang. 

\subsubsection{Solution Mutation}

Instead of building a SolFunc from scratch in Phase 1, sometimes the
Evolver picks an existing solution for another problem in the same
\emph{gang}, and attempts to mutate it by sending it through  Phases 1,
2 and 3. In AAD, solution mutation is a trade-off, because if we pick an
already built solution for mutation, we lose the opportunity to build a
fresh SolFunc in that step. Notice that solution-mutation is different
from operator-mutation described under Phase 3.

Natural evolution also takes advantage of multiple, related problems
present in an environment to test new solutions or adapt
existing ones for new purposes, as in the case of evolution of birds' feathers, which are
believed to have first served a thermoregulatory or a display
function~\cite[p.~671]{lifeOfBio}.

\subsubsection{Cross-Pollination Among Ranks}

\begin{figure}
  \includegraphics[width=0.45\textwidth]{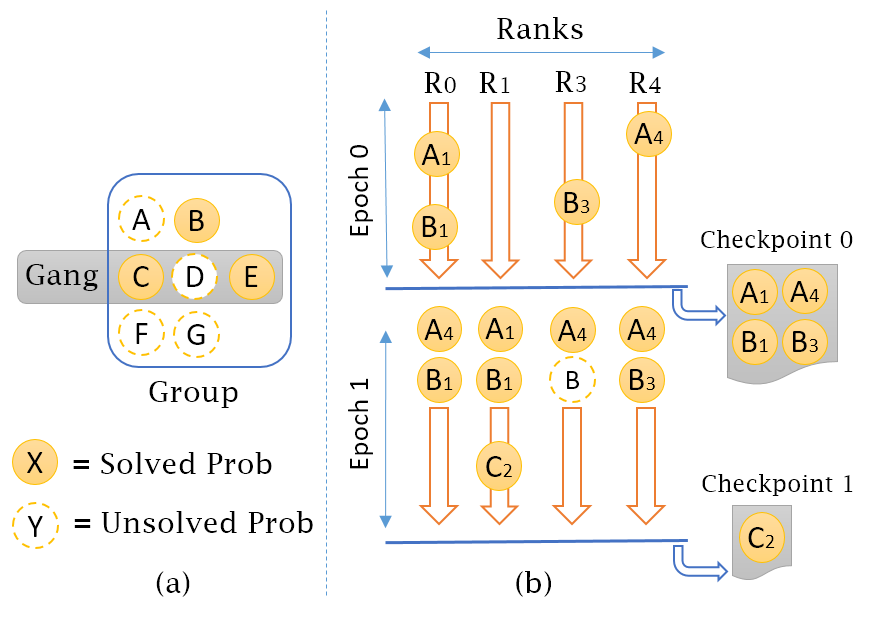}
  \small
  \caption{(a) How problems are grouped into {\em gangs} and {\em groups}.
           (b) Distribution of solutions via process joins and checkpoints. 
           A, B, C are problems, with the subscript indicating the rank on which
           the solution is found.}
  \label{fig:ranks}
  \normalsize
\end{figure}

AAD creates multiple concurrent processes (called {\bf ranks}) and
assigns problems to each of them to solve. Evolution happens in
multiple ranks in isolation, in periods called {\bf epochs}, as shown in 
Figure~\ref{fig:ranks}(b).
Pseudo-code for the high-level evolution algorithm for {\em one epoch} is shown in 
Figure~\ref{fig:singleEpoch}.

\codelistsize
\begin{figure}
\begin{verbatim}
function evolve_epoch(group, num_steps)
   foreach step in num_steps
      foreach gang in group
         foreach prob in gang
            if prob.solution == null
               evolve_one_step(prob, gang)

function evolve_one_step(prob, gang)
   SolFunc = new OR pick_existing_soln(gang)
   build(SolFunc)                         # Phase 1
   for num_link_attempts
      link_and_check(SolFunc, gang)       # Phase 2
      op_mutate_and_check(SolFunc, gang)  # Phase 3
      if check succeeded for any prob_x in gang
         add_func_call_to_ExpStore(SolFunc, prob_x)
\end{verbatim}
\caption{Evolution algorithm for single epoch.}
\label{fig:singleEpoch}
\end{figure}
\normalsize

At the end of an epoch, synchronization happens and solutions are exchanged among ranks,
as shown in Figure~\ref{fig:ranks}(b). At the end of an epoch, solutions
discovered by each rank are sent to the master rank, which collects and
distributes all of them to all ranks to be used in the next epoch. In
the next epoch, a rank may receive solutions discovered by any rank in
the previous epoch (see Figure~\ref{fig:ranks}(b)).

\subsubsection{Diverse Environments \& Solutions}
\label{design:diversity}

In AAD, each rank maintains its own copy of ExpStore and  AAD allows
some of the non-essential expressions in Table~\ref{tab:grammar} to be
randomly removed form a rank's ExpStore. In current setup, binary
\specname{Arithmetic} operations are added 80\% of the time to a rank,
{\em Pop at Ind} expressions are added 20\% of the time, and
Reduction idiom is added 10\% of the time. Further, when a {\em For
Stmt} is inserted, 20\% of the time we use an enumerated {\em For Stmt}
instead of a non-enumerated one. Moreover, when a rank receives a
solution to a problem, a function call for it is added to ExpStore only
80\% of the time. 
These random omissions of expressions, including
function calls, help create diverse environments with respect to
expressions available in the ExpStore.

Additionally, AAD allows multiple solutions (e.g., 100) to coexist for a
given problem. For a solved problem, a rank may receive any one of these
existing solutions, picked at random. This allows solution mutation to
start from different solutions.
Further, currently, 20\% of the ranks do not receive a solution for a
given problem, even when one exists, forcing them to find their own
solutions. This is illustrated in Figure~\ref{fig:ranks}(b) with an
empty circle, where Rank R3 does not receive a solution for problem B.
Both of these strategies increase diversity of solutions. 

These strategies are inspired by biological evolution, which uses
diverse environments resulting from different temperatures, salinity,
humidity, pressure, etc., to come up with different organisms. Similarly,
evolution depends on diversity of individuals (solutions) within a
population.

\subsection{Engineering Challenges}

Since AAD supports loops and conditional statements and uses function
composition, AAD presents several challenges that are not faced by
simpler program synthesizers. We take a practical engineering approach
to solving these challenges as outlined below.

\subsubsection{Exceptions}
Although AAD produces syntactically correct programs, many runtime
exceptions are possible due to various reasons ranging from divide by
zero to popping from an out-of-bound index. Fortunately, Python provides
a robust exception handling framework and AAD catches all exceptions
arising from Python's built-in \texttt{exec()} function used to execute
SolFunc. 

Allowing exceptions in the first place does not make AAD less robust. As
an engineering example, out-of-order processors introduce unintended
exceptions such as page faults, divide by zero exceptions, through
speculative actions taken by the processor
itself~\cite{speculative1,speculative2,speculative3}. However,
processors have mechanisms to detect such violations and correct
themselves, thereby making them robust.

\subsubsection{Infinite Loops}

Even natural evolution may cause infinite repetition, 
as seen with cancerous cell division.
Instead of trying to detect infinite repetion (loops), we use a timeout to terminate
programs that do not terminate within a given time period (e.g., 1
second), using Python's \emph{signal} module to set an ALRM signal. Although it
may appear as a crude solution, this is a well established engineering
technique used in complex systems like microprocessors and spacecrafts, which use
various watchdog timers~\cite{watchdog2,watchdog3,watchdog4} to recover from a multitude of
unpredictable situations like deadlocks, live-locks (starvation), soft
errors, etc. For instance, when an execution core of a microprocessor 
issues a memory read request, if it gets dropped in the memory system due
to an unexpected condition, the core may timeout and reissue the
request. 

However, there is a cost to this approach: timeouts waste valuable
processing time. Therefore, we use one heuristic to prevent one such
cause of infinite loops. When we iterate over an array $A$ using a
\specname{for loop}, allowing $A$ to grow by appending more items can
cause an infinite loop. In Python, this can be easily prevented by
making $A$ a tuple, which is immutable.

\subsubsection{Infinite Recursion}

Since cycles in call graph can lead to infinite recursion, AAD disallows
any recursion. Since recursion detection is well understood~\cite{recursion1,recursion2},
AAD detects and discards programs with recursion. When we construct SolFunc, we do
not let SolFunc call itself. When we call an already built function $F$
within SolFunc, we do not allow $F$ to call SolFunc either. Therefore,
we do not have a cycle between SolFunc and $F$. We repeat this cycle
detection process for any function $F$ we use.

\subsection{Parallelization and HPC}
The parallelization of the evolutionary search process naturally lends
to High-performance Computing {\bf (HPC)}, where a large number of
processors and nodes are used to solve a problem. Therefore, we designed
AAD as a multiprocess application, using Python's multiprocessor module,
to take advantage of multiple processors on a single node. In addition,
as shown in Figure~\ref{fig:ranks}(b), AAD can take a checkpoint of
solutions discovered since the last checkpoint. With distributed file
systems available on HPC clusters, this allows solutions found on node
$N$ to be exchanged with other nodes, by those nodes reading the
checkpoint dumped by node $N$. A node can read any available checkpoint
at the start of an epoch, and does not wait for any specific checkpoint,
thereby avoiding any synchronization. A checkpoint is produced using
Python's {\em pickle} module.

However, parallel processing introduces other engineering challenges,
including load balancing. Due to diverse environments and random nature
of evolutionary decisions, some ranks may take longer to finish an epoch
than others. Similarly, a large number of timeouts due to infinite
loops can also increase the execution time of a rank. To mitigate the
latter problem, a rank counts the number of total timeouts and if it is
greater than a threshold, ends epoch early. 

Strategies like discarding of solutions and early termination of search
are made possible by the non-deterministic evolutionary nature
of AAD. Evolution does not put much value on a single individual or even
a single environment, but mainly depends on the continuation of the
process itself. As long as evolution continues, a solution may be found
one way or the other.

\section{Methodology}
\label{aad:methodology}

The entire AAD framework is written in Python and is about 6,700 lines of
code (see Appendix E), including all problem generators and checkers,
blank lines and comments.

We use 3 groups of 10 problems each (described in Appendix A) to study the effectiveness of
evolutionary strategies described in Section~\ref{design:strategies}.
All problems are listed in Section~\ref{results:composition}.
GroupA consists of typical list manipulation problems (e.g., min, sort,
reverse, index, remove), GroupB represents basic vector processing
problems (e.g., dot-product, matrix-vector multiply), and GroupC
consists of some basic spreadsheet problems (e.g., sum, sum-if,
count-if, average). 
One entire group is evolved in a given run. We run
112 concurrent processes on a 4-socket Intel Xeon(R) 8180 Platinum server
with a total of 112 physical cores. To minimize artifacts due to
randomness, we do 10 runs per experiment, for each group.

The metric used to evaluate different strategies is {\em steps}. Within
a single step, we can send a SolFunc through phases 2 and 3 any number
of times, to re-link and re-mutate. Currently, we do this twice, leading
to 4 different variants. Since we simulate 112 ranks concurrently, each
step reported accounts for 448 $(112*4)$ distinct SolFuncs. On the above
server, on average, one step on a rank takes about 7 ms per gang.
Runtime, however, is {\em not} an objective of this paper and we make no
claims regarding runtime nor make any attempts to compare runtime of AAD
to prior work. 

For each problem, when a solution is found on a rank, it is reported by
writing the solution and relevant statistics to a log file created for
each rank. The first solution to be found for a problem among any rank
is found through post-processing log files and its step count is
reported as the time to solution for that problem for that run. For an
experiment with 10 runs, for a given problem, we report 
the average of all such step counts across all 10 runs.

For each run, we simulate a maximum of 100 epochs, with each epoch
containing 2,000 steps. If at least one solution is found for all
problems in a group before 100 epochs, we end the run and dump a checkpoint. 
For reporting purposes, AAD can {\em rank} solutions by reading  checkpoints
from one or more runs to find the {\em least complex} ones.
Although there are many strategies to 
rank~\cite{ranking1,ranking2,ranking3}, we use a simple heuristic
that assigns different weights to different structures -- e.g., 50 for
a function call, 20 for a \specname{For Stmt}, 10 for an \specname{If
Stmt}, and 2 for any other statement.
Ranking for all other purposes is left as future work.

AAD supports different parameters. Currently, we allow a maximum of 12
statements for a SolFunc, with additional 2 statements if SolFunc is
mutated from an existing solution. We allow up to 100 different
solutions to be found for a single problem. Default parameters used to
create diverse environments were given in
Section~\ref{design:diversity}. All values were picked as reasonable
guesses and tuning them is left as a future study.

There are no restrictions placed on inputs of problems other than all
arrays must be non-empty. Problem generators usually generate input
arrays from size 1 to 200, randomly filled with integers from -200 to
+200.

\section{Results and Analysis}

In this section, we present results showing the
effectiveness of different evolutionary strategies
discussed in Section~\ref{design:strategies},
along with insights from code generated by AAD.

\label{aad:results}

\subsection{Evolutionary Strategies}

\begin{table}
\small{
\begin{tabular}{l| c c c c c c c c c c} 
        & \rotatebox{90}{ Max}  & \rotatebox{90}{ Min}  & \rotatebox{90}{ SortDesc}     & \rotatebox{90}{ SortAsc}      & \rotatebox{90}{ ReverseArr} & \rotatebox{90}{ RemoveL}      & \rotatebox{90}{ RemoveF}      & \rotatebox{90}{ LastIndOf}    & \rotatebox{90}{ FirstIndOf}   & \rotatebox{90}{ IsInArr}    \\ \hline
 Max    &       &       &       &       &       &       &       &       &       &       \\
 Min    &       &       &       &       &       &       &       &       &       &       \\
 SortDesc       &57     &14     &       &43     &43     &14     &57     &       &       &       \\
 SortAsc        &4      &8      &94     &       &94     &6      &8      &       &       &18     \\
 ReverseArr     &       &       &       &       &       &       &       &       &       &       \\
 RemoveL        &2      &1      &       &       &26     &       &22     &79     &       &10     \\
 RemoveF        &1      &1      &       &       &59     &61     &       &       &43     &14     \\
 LastIndOf      &       &2      &       &       &       &       &       &       &       &       \\
 FirstIndOf     &       &       &       &       &       &       &2      &       &       &       \\
 IsInArr        &2      &3      &       &       &3      &1      &18     &61     &3      &       \\ 
\end{tabular}
\caption{Caller-callee relationships showing function composition for GroupA problems.}
\label{tab:callees}
}
\vspace{-0.1in}
\end{table}

\subsubsection{Composition} 
\label{results:composition}

Table~\ref{tab:callees} shows caller-callee
relationships for GroupA problems (see Appendix B for all groups). 
For instance, the row for \texttt{SortDesc}
function shows that \texttt{SortAsc} called \texttt{Max} in 57\% of the solutions, \texttt{Min} in
14\% of the solutions, and so on. Functions \texttt{Min}, \texttt{Max}, and \texttt{ReverseArr} did
not call any other function. All other functions depended on one or more
functions to arrive at a solution, underscoring the importance of
function composition. Some of the function dependencies may be
unexpected as discussed in Section~\ref{sec:out-of-box}. 

\begin{table}
\small{
\begin{tabular}{|l|r||r|r|r|r|} \hline
{\bf GroupA} &        {\bf Baseline} &     {\bf Exp1} &  {\bf Exp2} &  {\bf Exp3} & {\bf Exp4} \\         \hline 
Max &           98 &    0.9 &   1.1 &   0.9 &   1.1 \\
Min &           90 &    0.8 &   2.6 &   0.9 &   1.0 \\
SortDesc &      59929 &         $\infty$ &      $\infty$ &      $\infty$ &      1.2 \\
SortAsc &       60177 &         $\infty$ &      $\infty$ &      $\infty$ &      1.2 \\
ReverseArr &    160 &   0.9 &   1185.9 &        1.2 &   1.5 \\
RemoveL &       2599 &  $\infty$ &      $\infty$ &      3.1 &   0.8 \\
RemoveF &       4050 &  $\infty$ &      $\infty$ &      5.2 &   0.8 \\
LastIndOf &     511 &   1.3 &   $\infty$ &      0.7 &   2.1 \\
FirstIndOf &    17077 &         1.8 &   $\infty$ &      2.5 &   0.7 \\ 
IsInArr &       269 &   0.8 &   $\infty$ &      0.9 &   1.4 \\  \hline \hline
{\bf GroupB} &         &       &       &       &       \\     \hline
AddArrays &     2349 &  1.0 &   0.7 &   2.7 &   1.3 \\
MultArrays &    18797 &         0.3 &   10.6 &  7.5 &   3.1 \\
Sum &           9 &     2.6 &   0.9 &   1.1 &   1.1 \\
SumOfSq &       9373 &  7.9 &   $\infty$ &      11.0 &  1.8 \\
DotProd &       18290 &         $\infty$ &      $\infty$ &      7.8 &   2.9 \\
MatVecMult &    54476 &         $\infty$ &      $\infty$ &      $\infty$ &      2.1 \\
AddToArr &      232 &   0.7 &   0.8 &   1.8 &   1.8 \\
SubFromArr &    2995 &  1.9 &   $\infty$ &      4.5 &   0.7 \\
ScaleArr &      3682 &  0.5 &   $\infty$ &      2.7 &   0.7 \\ 
ScaledSum &     29 &    269.0 &         $\infty$ &      2.0 &   0.8 \\  \hline \hline
{\bf GroupC} &         &       &       &       &       \\     \hline 
CountEQ &       91345 &         $\infty$ &      $\infty$ &      1.5 &   1.9 \\
CountLT &       109444 &        $\infty$ &      1.2 &   1.3 &   1.6 \\
CountGT &       109431 &        $\infty$ &      1.2 &   1.3 &   1.6 \\
SumIfLT &       25109 &         5.7 &   5.7 &   5.4 &   1.7 \\
SumIfGT &       25065 &         5.7 &   5.7 &   4.6 &   1.7 \\
SumIfEQ &       25001 &         5.7 &   8.0 &   4.5 &   1.7 \\
ScaledAvg &     7548 &  $\infty$ &      $\infty$ &      1.1 &   0.8 \\
Sum      &      7 &     18.6 &  0.6 &   1.6 &   2.3 \\
Len      &      28 &    0.5 &   1.1 &   0.5 &   0.2 \\
Avg      &      891 &   $\infty$ &      $\infty$ &      1.2 &   1.2 \\  \hline
\end{tabular}
\caption{Effectiveness of different evolutionary strategies. 
Exp1: Composition, 
Exp2: Operator mutation, 
Exp3: Cross-pollination among ranks,
Exp4: Diversity of solutions.}
\label{tab:effectiveness}
}
\vspace{-0.3in}
\end{table}

Table~\ref{tab:effectiveness} shows all three groups of problems, their
baseline performance in terms of the step count, and compares the
effectiveness of four evolutionary strategies against the baseline. For
the four experiments shown, each entry gives the ratio between the
number of steps taken without that strategy and that of the baseline.
Baseline step counts span a wide range due to the varying complexity of problems.
Since step counts for each entry is taken by averaging across multiple
runs, when a run does not produce a solution for a given problem, the
maximum step count (200,000) is used for that run~\cite{pldi18}. If none
of the runs for a given experiment produces a solution for a given
problem, symbol $\infty$ is used to indicate that fact.

Step counts for Exp1 are obtained by disabling the addition of function calls
to ExpStore for solutions found (i.e., disabling composition). This
increases time to solution by a significant factor for some problems
(e.g., \texttt{SumOfSq}, \texttt{ScaledSum}) and makes it impossible to
find any solutions for others (e.g., \texttt{SortAsc},
\texttt{DotProd}), underscoring the importance of composition,
especially for more complex problems. However, not adding function calls
to ExpStore shrinks the search space and hence can speed up solutions
of some simpler problems (e.g, \texttt{Len}, \texttt{IsInArr}) that do
not have to depend on other functions.

\subsubsection{Operator Mutation} 
Exp2 disables operator-mutation by disabling Phase 3, leading to
increased solution times and altogether unsolved problems, especially in
GroupA. Solutions that depend on an alternative operation (e.g.,
multiply instead of addition) are susceptible to eliminating
operator-mutation. However, having fewer (or late) solutions reduces
search space by reducing function composition and hence can speed up
solutions to problems that do not depend on operator-mutation (e.g.,
\texttt{Sum}).

\subsubsection{Cross-Pollination}
Exp3 in Table~\ref{tab:effectiveness} disables cross-pollination among
ranks by running a single epoch of 200,000 steps, thereby avoiding any
exchange of solutions among ranks. Many problems that depend on others
(e.g., \texttt{SortAsc}, \texttt{SortDesc}) are severely affected by lack of
cross-pollination. 

\subsubsection{Diversity}
Exp4 reduces the number of solutions maintained for a given problem from
100 to 1, thereby decreasing the {\em diversity of solutions} for a
given problem. This negatively affects several problems (e.g.,
\texttt{MultArrays}, \texttt{DotProd}) demonstrating the importance of solution-diversity.
Similarly, for GroupC, Exp5 in Table~\ref{tab:soln-mutation} shows the effect of adding all optional expressions
described in Section~\ref{design:diversity} 
100\% of the time, decreasing {\em diversity of environments} and
increasing expressions available to every rank. This shows that for many
problems in GroupC adding optional expressions randomly is a better
choice than always adding them to every rank. However, as before,
simpler problems (e.g., \texttt{Len}) can benefit from the late
discovery of other solutions.

\begin{table}
\small{
\begin{tabular}{l| c c c c c c c c c c}
        & \rotatebox{90}{CountEQ}       & \rotatebox{90}{CountLT}       & \rotatebox{90}{CountGT}       & \rotatebox{90}{SumIfLT}       & \rotatebox{90}{SumIfGT}       & \rotatebox{90}{SumIfEQ}       & \rotatebox{90}{ScaledAvg}     & \rotatebox{90}{Sum}   & \rotatebox{90}{Len}   & \rotatebox{90}{Avg}   \\ \hline
CountEQ &       &38     &33     &1      &6      &19     &       &       &       &       \\
CountLT &43     &       &33     &2      &3      &16     &       &       &       &       \\
CountGT &43     &36     &       &1      &3      &15     &       &       &       &       \\
SumIfLT &2      &       &       &       &41     &56     &       &       &       &       \\
SumIfGT &1      &1      &       &43     &       &53     &1      &       &       &       \\
SumIfEQ &       &1      &       &42     &53     &       &1      &       &       &       \\
ScaledAvg       &       &       &       &22     &25     &34     &       &       &       &       \\
Sum     &       &       &       &       &       &       &       &       &       &       \\
Len     &       &       &       &       &       &       &       &5      &       &       \\
Avg     &       &       &       &       &       &       &       &       &       &       \\
\end{tabular}
\caption{\% of solutions where a solution  mutated from another solution (a parent), for GroupC problems.}
\label{tab:parents}
}
\vspace{-0.25in}
\end{table}

\subsubsection{Solution Mutation} 
Table~\ref{tab:parents} gives \% of solutions
where a solution is mutated from another solution (a parent) for GroupC
problems, where this is most common (see Appendix C for all groups). 
It should be noted that mutation can happen in either
direction, on different ranks. For instance, some ranks may first come
up with \texttt{Min} (or receive \texttt{Min} as an already solved
problem from another rank) and \texttt{Max} may mutate out of it. On
other ranks, \texttt{Min} may mutate out of \texttt{Max}.
\small
\begin{table}
\begin{tabular}{|l|c c c c c c c c c c|} \hline

  &
\rotatebox{90}{CountEQ}       & \rotatebox{90}{CountLT}       & \rotatebox{90}{CountGT}       & \rotatebox{90}{SumIfLT}       &\
\rotatebox{90}{SumIfGT}       & \rotatebox{90}{SumIfEQ}       & \rotatebox{90}{ScaledAvg}     & \rotatebox{90}{Sum}   & \rotatebox{90}{L\
en}   & \rotatebox{90}{Avg}   \\ \hline

Exp5 &1.8 & 1.4 & 1.4 & 2.0 & 2.0 & 2.0 & 2.1 & 1.3 & 0.4 & 1.5\\ 
Exp6 & 2.0 & 1.6 & 1.6 & 3.7 & 5.8 & 2.2 & 0.9 & 0.6 & 0.4 & 0.9\\ \hline 
\end{tabular}
\caption{Effectiveness of diversity of environments (Exp5), solution mutation (Exp6).}
\label{tab:soln-mutation}
\vspace{-0.25in}
\end{table}
\normalsize

Exp6 in Table~\ref{tab:soln-mutation} shows the effects of disabling
solution-mutation for GroupC problems, which are dependent on parents as
given by Table~\ref{tab:parents}. Problems that mutate from parents show
increased time to solution, while problems that do not (e.g.,
\texttt{Len}) see a speedup, as in previous experiments. 

\small
\begin{table}
\begin{tabular}{|l| c c |c|} \hline
         & AddArrays   & MultArrays & Total \\ \hline
Separate & 2023        &   7866     &   9889 \\
Together & 1746        &   6105     &   6105 \\ \hline
\end{tabular}
\caption{Exp7: Effectiveness of ganged-evolution.}
\label{tab:ganged}
\vspace{-0.25in}
\end{table}
\normalsize

\subsubsection{Ganged Evolution} 
To show the effectiveness of ganged-evolution,
we picked two problems that belong to the same gang, evolved them one at
a time, and compared the results to evolving them together, as shown
with Exp7 in Table~\ref{tab:ganged}. To isolate effects of ganged-evolution, 
we disabled function composition and solution-mutation, and
picked two problems that can evolve without other functions. Results
show that evolving both of them separately takes about 1.6X steps
(9,889) compared to evolving them together (6,105). \emph{Total} column
reports total steps to find solutions for both problems, in each case.

It should be emphasized that every evolutionary strategy is not important for every problem. Some
simple problems can directly evolve from the grammar itself and they are
often hurt by advanced strategies used. However, as these results show,
many complex problems cannot find a solution without these strategies,
within a reasonable time limit.

\subsection{Code Examples and Insights}

Being able to generate complex code is an important result of AAD.
Besides \texttt{SortAsc} (from GroupA) shown in Section~\ref{aad:introduction}, 
in this section we show an example each from GroupB and GroupC. 
Code generated for all 29 problems is shown in Appendix D.
Since AAD finds multiple solutions for a problem, we only discuss one
of them, usually the least complex one.

\begin{figure}
\codelistsize
\begin{verbatim}
def Sum(arg0):
        arr_10 = arg0.copy()
        num_14 = 0
        for num_12 in tuple(arr_10):
                num_14 = num_12 + num_14
        return (num_14)

def MultArrays(arg0, arg1):
        arr_10 = arg0.copy()
        arr_11 = arg1.copy()
        for num_13 in tuple(arr_10):
                num_14 = arr_11.pop(0)
                num_16 = arr_10.pop(0)
                num_15 = num_16 * num_14
                arr_11.append(num_15)
        return (arr_11)

def DotProd(arg0, arg1):
        arr_10 = arg0.copy()
        arr_11 = arg1.copy()
        arr_14 = MultArrays(arr_10, arr_11)
        num_13 = Sum(arr_14)
        return (num_13)

def MatVecMult(arg0, arg1):
        arr_of_arr10 = arg0
        arr_11 = arg1.copy()
        arr_16 = list()
        for arr_15 in tuple(arr_of_arr10):
                num_17 = DotProd(arr_11, arr_15)
                arr_16.append(num_17)
        arr_14 = arr_16.copy()
        return (arr_14)
\end{verbatim}
\caption{Code for \texttt{MatVecMult}}
\label{lab:listing}
\normalsize
\end{figure}

The solution for \texttt{MatVecMult} (Figure~\ref{lab:listing}) is performing dot products
(\texttt{DotProd}) between row vectors in the matrix (arg0) and the
input vector (arg1), and appending those results to a new result vector.
\texttt{DotProd} in turn depends on the sum (\texttt{Sum}) of two arrays
multiplied together (\texttt{MultArrays}). \texttt{MatVecMult} is
performing a linear transformation, which is the basis of linear
algebra, and hence this discovery of AAD is particularly noteworthy. 

From GroupC, the solution for \texttt{CountEQ}, which counts the number of times
an element occurs in an array, is shown below. The algorithm is somewhat circuitous, because
first it appends matching elements to a new list and then finds the length
of that list using \texttt{Len}. This is an example of a non-obvious algorithm, although it is not
the most efficient solution.

\codelistsize
\begin{verbatim}
def Len(arg0):
        arr_10 = arg0.copy()
        num_15 = 0
        num_14 = 1
        for num_12 in tuple(arr_10):
                num_15 = num_14 + num_15
        return (num_15)

def CountEQ(arg0, arg1):
        arr_10 = arg0.copy()
        num_11 = arg1
        arr_14 = list()
        for num_15 in tuple(arr_10):
                bool_17 = (num_11 == num_15)
                if (bool_17):
                        arr_14.append(num_15)
        arr_10.append(num_15)
        num_12 = Len(arr_14)
        return (num_12)
\end{verbatim}
\normalsize

It should be noted that the code generated by AAD often contains
redundant operations like copying, appending and popping the same element,
calling functions whose results are never used, if-statements with
always True or False conditions, etc. Some of these
can be easily eliminated with standard compiler techniques and
are out of scope for the current version of AAD.

\subsubsection{Outside-of-the-box Solutions}
\label{sec:out-of-box}

This section describes some unexpected programs illustrating both
strengths and weaknesses of AAD. Since AAD uses a limited number of
input combinations generated by a ProbGen to check the validity of a
program, any solution generated for a problem is as good as the ProbGen
and the Checker used. This is both a weakness and a strength depending
on the application. If an application demands rigorous validation, it is
the responsibility of the ProbGen and the Checker to cover all cases,
including corner cases. 
Writing such verification logic can be quite demanding, which is an
obvious weakness. On the other hand, when an application needs to take
advantage of peculiarities of the input, or needs to come up with
solutions in constrained environments, AAD can show remarkable
adaptability.

As a simple example, the grammar we use does not have truth values True
and False provided as constants. Although initially this was an omission
on our part, we realized that AAD was actually generating these values
when they were needed, using an expression of the form \texttt{bool\_10
= (num\_11 == num\_11)}. Similarly, initially, we forgot to include
constant values 0 and 1 in the grammar. AAD overcame that difficulty by
subtracting the same value from itself to generate zero and dividing the
same value (e.g., the last value of  an array) by itself to to generate
constant 1. Although, the latter approach is not safe because the last
value of an array could be zero, it used that in cases where the ProbGen
did not put a zero at the end of an array. To defeat this, we changed
ProbGen to generate an array of all zeros for some problem sizes. Then
AAD generated constant value 1 by taking advantage of \texttt{Len} -- by
dividing the length of the array by itself, because we always use
non-empty arrays. This shows that the SolGen is in an adversarial
relationship with ProbGen/Checker, trying to defeat the latter duo by
exploiting any opportunity or weakness present, similar to bacteria
adapting to anti-biotics in biolgical evolution.

Another such example is \texttt{FirstIndOf}, which is used for finding
the index of a given element in an array. If there are multiple elements
present, the function returns the index of the first element found. Most
programmers would write a for-loop to iterate over the array looking for
the element, and when a match is found, will break out of the loop using
a break or a return statement. However, the current grammar used by AAD
does not have a break statement and does not use return statements in
the middle of a function. One solution AAD came up with to solve this
challenge is given below. First, it goes through the array popping each
element from the front. If a match is found, it is appended at the end
of the same array. Once it has gone through the loop, all the elements
left in the array are the matching elements appended at the end. Then
this function returns the head of the remaining array, which is the
first matching index. In other situations, we have also seen it calling
\texttt{Min} function to find the minimum value of the remaining array.

\codelistsize
\begin{verbatim}
def FirstIndOf(arg0, arg1):
        arr_10 = arg0.copy()
        num_11 = arg1
        for num_12, num_13 in enumerate(tuple(arr_10)):
                num_17 = arr_10.pop(0)
                bool_16 = (num_17 == num_11)
                if (bool_16):
                        arr_10.append(num_12)
                num_13 = arr_10[0]
        return (num_13)
\end{verbatim}
\normalsize

\texttt{FirstIndOf} is usually defined with a pre-condition -- viz., the
element must be always present (e.g., \texttt{index()} in Python). With
our ProbGen/Checker, the behavior of this function is not defined if
arg1 is not present in the array. We observed \texttt{IsInArr} function
exploiting this undefined return value of a specific \texttt{FirstIndOf}
function to arrive at a solution, which always passed our checker but
was incorrect in a rare situation the Checker did not check for.
Although such a solution is problematic in many uses, it illustrates the
ability to take advantage of deficiencies (or features) present in the
inputs or the Checker. This could be quite valuable for an autonomous
system to adapt to (or take advantage of) vulnerabilities of an
adversary. 

The following solution for \texttt{IsInArr} is a prime example of
outside-of-the-box thinking. This code finds whether a given element
(arg1) is in a given array (arg0). Most programmers would write a loop
that goes over the array and looks for a match. However, the following
program does not contain any loops, which appeared to be a bug
until we realized what it was doing. 
It first appends arg1 to the end of
the array. Then, it calls \texttt{RemoveF} to remove arg1, which removes
the first matching element. Then it pops the last element of the array
and checks whether it matches arg1 it appended. If it matches arg1,
\texttt{removeF} must have removed another element equal to arg1 from the array.
In that case, there must have been another element already present in
the array and \texttt{IsInArr} must return true, as it does here. If the
popped element does not match arg1, it means \texttt{RemoveF} removed arg1 it
appended itself. In that case, there was no prior matching element and
\texttt{IsInArr} must return False, as is the case.

\codelistsize
\begin{verbatim}
def IsInArr(arg0, arg1):
        arr_10 = arg0.copy()
        num_11 = arg1
        arr_10.append(num_11)
        arr_10 = RemoveF(arr_10, num_11)
        arr_18 = arr_10.copy()
        num_13 = arr_18.pop()
        bool_12 = (num_13 == num_11)
        return (bool_12)
\end{verbatim}
\normalsize

If any person were to come up with this solution for \texttt{IsInArr}, we would
have labeled him or her as \emph{creative} or an \emph{outside-of-the-box}
thinker. 
It is hard to 
believe that a machine is capable of this level of \emph{logical reasoning}, 
although AAD stumbled upon it without any reasoning at all. 
This capability presents us with a new opportunity for AAD -- viz. to
use it as an \emph{outside-of-the-box thinker} to come up with
alternative solutions that we would not normally think of. After all,
creative thinking does not seem to be the prerogative of
humans alone.

\section{Related Work}
\label{aad:relatedwork}

Program synthesis is an active and challenging research area that seeks
to automate programming.
Many approaches have been proposed over the years in an effort to
generate programs that conform as accurately as possible to the
user-expressed intent. In~\cite{progSynthNow} Gulwani, Polozov, and
Singh present an excellent survey on the program synthesis problem,
applications, principles and proposed solutions. In~\cite{progSynth18},
the same authors extend their prior survey to include more
recent advances that span 2017 and 2018. Below, we review related work
in the area, discuss typical problem domains, challenges in program
synthesis and position our work among prior research.

Program synthesis approaches have targeted automatic generation of
solutions for problems in domains such as data
wrangling~\cite{flashFill,SinghNumTran,SinghSpread,flashExtract},
graphics~\cite{graph1,graph2}, code
repair~\cite{codeRep1,codeRep2,codeRep3},
superoptimization~\cite{superOptThai,superOpt2,superOpt3,superOpt4}, and
others. In the above cases solutions are sought for restricted target
problem domains such as string manipulation~\cite{flashFill}, bit-vector
programs~\cite{bitvector1}, optimized code implementations at the ISA
level~\cite{superOptThai}, etc. Moreover, many program synthesis
approaches are restricted to straight-line code
fragments~\cite{superOptThai,superOpt4,deepCoder}. While there exist
works that target loop-based code, they are either restricted in the
form of SIMD code~\cite{loopSIMD} or synthesis of loop bodies (e.g.,
within templates~\cite{Srivastava2013} or sketches~\cite{sketchLoop});
the majority of related work, unlike AAD, focus on loop-free 
programs~\cite{bitvector1,loopFree}.
However, program synthesis tools like SYNQUID~\cite{types1}, 
MYTH~\cite{types2}, and Leon~\cite{leon} can generate recursive and/or
higher-order programs that can be as expressive as loop-enabled approaches,
in frameworks with formal specifications. 
AAD enables similar capabilities for general-purpose Python language 
with its support of loops, function composition, and complex control flow.

Two of the main challenges in program synthesis are related to the
intractability of program search space and accurately expressing user
intent.
There are many search techniques proposed to address the former problem:
enumerative, deduction-based~\cite{flashMeta}, constraint
solving~\cite{SolarThesis,Srivastava2013}, as well as stochastic search
techniques~\cite{MLPBE,Koza1994}. Stochastic search techniques include a lot of modern
approaches that employ machine learning~\cite{MLPBE,ML2} and neural
program synthesis~\cite{deepCoder,neural1,neural2,neural3}. 
In comparison, AAD uses a modified evolutionary approach that
relies on PGE without requiring a fitness function.

As far as expressing user intent is concerned, different program synthesis
techniques use formal logical specifications (mostly deductive-based
techniques), informal natural language descriptions, and templates and
sketches, among others. In AAD we provide the specification in the form
of a program (called a \emph{checker}) along with test inputs. This is
similar to oracle-guided synthesis~\cite{bitvector1} (where an oracle produces
the correct output) and reference implementations in SKETCH~\cite{SolarThesis}.

To show the effectiveness of AAD, we use array based problems 
similar to the benchmarks (e.g., \emph{array-search}) in the
SyGuS-Comp competition~\cite{sygusComp}. However, solvers such as Sketch-based,
enumerative or stochastic, do not scale up to large array
sizes~\cite{sygusPaper}, although it may be conceptually possible to extend
Sketch based templates to express the grammar we use, in order to support large arrays.
AAD’s support of loops allows it to support input arrays of any (non-zero) size, 
and hence be as effective as frameworks that support recursion and/or higher-order operators. 

Overall, our work complements and builds upon prior approaches
that use composition; Bladek 
and Krawiec~\cite{bladek} propose a similar genetic programming approach 
of simultaneous synthesis of multiple functions, and briefly explain the
concept with four simple examples (last, patch, splice, splitAt). AAD extends 
this to much more complex
problems with PGE and associated evolutionary strategies.
Although, other works, like SYNQUID~\cite{types1} utilize
\emph{components} in the process of program synthesis, unlike the former,
PGE does not require the user to specify any underlying 
order (dependence) between the constituent components/functions. 
Besides, not specifying dependencies allows AAD to discover 
{\em outside-of-the-box} solutions. 

Compared to other works (e.g., 
deductive-based approaches), program equivalence in AAD is not formally 
proven. This is typical in similar approaches of counter-example guided 
synthesis where "the best validation oracle for most domains and 
specifications tends to be the user"~\cite{progSynthNow}, who can inspect 
the program under consideration. 
We also emphasize that formal verification is not a prerequisite for many 
useful applications, especially for {\em knowledge discovery} for AI (e.g.,
for a robot to find a way to sort objects for packing). 
Human knowledge in general is inductive in nature. 
After all, biological evolution produced complex and intelligent organisms as 
humans without anyone writing a formal specification.

\section{Discussion \& Future Work}
\label{aad:discussion}

This section discusses limitations of AAD, alternative ways of guiding evolution,
and potential applications of AAD.

\subsection{Limitations of AAD}
A large search space is a challenge to search-based
synthesizers~\cite{progSynthNow}. For AAD, this is especially true due
to addition of function calls as new solutions are discovered.
AAD depends on guiding to solve this challenge and 
Section~\ref{disc:guiding} outlines several possible ways for guiding.

The problems solved in this paper mostly require regular control flow
(except \texttt{FirstIndOf} and \texttt{RemoveF}). Programs that require complicated
control-flow may take considerably more time to be discovered.
Similarly, algorithms that depend on a very specific value (e.g.,
\texttt{if x $<$ 3.5}) are hard for AAD to discover, unless those values
are present. AAD would be more suitable for performing permutations and
combinations of already available inputs, with straightforward numerical
processing. The solution for the above deficiency is to have proper
library support. For instance, although it may be difficult for AAD to
produce an algorithm for FFT (Fast Fourier Transform) on its own, it 
should be able to call an FFT implementation in a library and use that 
to solve other problems.

\subsection{Guiding the Hand of Evolution}
\label{disc:guiding}

Grouping problems for PGE can be achived in several ways. 
First, we can imagine humans (domain experts) doing guiding.
For instance, future `programmers' or scientists could just suggest AAD
to use problems A and B to come up with a solution for problem C (e.g.,
``try using dot product to come up with an algorithm for matrix-vector
multiply''). Notice that this is quite analogous to the way we teach
children to discover solutions to problems on their own (``try using a
screwdriver instead of a hammer''). Similarly, a
researcher who wants to come up with a hypothesis, or a programmer who
wants to come up with a heuristic, may be able to make some suggestions
and let AAD discover an algorithm, especially a non-obvious one, based
on that guidance. If the Checker is based on past data or sensor data
from the physical environment, this strategy could be used on many
real-world problems without having to write a Checker or a ProbGen as we
discussed in Section~\ref{sec:checker}. This would be an entirely new
way to ``program'' computers and build scientific models, and we intend
to pursue this further. 

Second, we can imagine other AI programs doing this guiding, especially
in restricted domains where AI systems can guess the components of a
solution based on domain, but not the exact algorithm~\cite{deepCoder}.

\subsection{Other Potential Applications}

Conceptually, AAD can also be used in program translation. If we have a
routine written in C, assembly, or even binary language, we can execute
that routine as a Checker for AAD to produce code in Python (or
similar). This is akin to a machine learning an algorithm by just
observing how another one behaves (i.e., how another one responds to inputs).
Incidently, Python code shown in this paper can be considered as Python to Python
translations, since the Checker is a different Python implementation. 

AAD could be more than a program synthesizer. It could be used to
acquire {\em intrinsic knowledge} for machines. The caller-callee graph
(Table~\ref{tab:callees}) and the parent-child graph
(Table~\ref{tab:parents}) capture inherent relationships between
different problems. For instance, we can see min/max is related to
sorting, and dot-product to matrix-vector multiply. These relationships
are discovered by AAD itself and can be thought of as one representation
of associative memory among {\em actions}, similar to what human brains
construct (e.g., getting ready in the morning is associated with
brushing teeth, dressing up, etc.).
Since AAD allows incremental expansion of knowledge by introducing more
and more problems, with a proper guiding mechanism we may be able to
guide autonomous systems to acquire a large number of skills
(algorithms) and build a knowledge representation on their own, the same
way we guide our children to acquire a large body of skills and
knowledge by presenting them with many problems and challenges over
their childhood.

\section{Conclusion}
\label{aad:conclusion}

We presented AAD, an evolutionary framework for synthesizing programs of
high complexity. Using a basic subset of Python language as
grammar, AAD allowed us to synthesize code for 29 array/vector problems, ranging from
min, max, reverse to more challenging problems
like sorting and matrix-vector multiplication, 
without input size restrictions. AAD's use of {\em problem guided evolution} (PGE) and related
evolutionary strategies made this possible. We
evaluated the effectiveness of these strategies and presented evidence
of \emph{outside-of-the-box} problem solving skills of AAD. To deal
with various challenges posed by complex requirements, 
we demonstrated how to use HPC techniques. Overall, we show that 
evolutionary algorithms with PGE are capable of solving problems of
similar or higher complexity compared to the state-of-the-art.

\bibliography{references}
\clearpage
\pagenumbering{arabic}
\appendix
\setcounter{figure}{0}
\setcounter{table}{0}
\renewcommand*{\thesubsection}{\arabic{subsection}}

\section*{Appendix A: Problem Definitions}
\label{aad:appendixA}

Table~\ref{tab:fullproblems} lists all 3 groups of problems used in this work.

\begin{table}[h]
\footnotesize{
\begin{tabular}{l  p{2.5in} }

{\bf Problem}    & {\bf Description} \\ \hline

{\bf GroupA} & \\ \hline
Max    & {\em NUM} = \texttt{Max}(ARR). Returns maximum of an array \\
Min    & {\em NUM} = \texttt{Min}(ARR). Returns minimum of an array \\
SortDesc & {\em ARR} = \texttt{SortDesc}(ARR). Returns a sorted array in descending order \\
SortAsc & {\em ARR} = \texttt{SortAsc}(ARR). Returns a sorted array in ascending order \\
ReverseArr & {\em ARR} = \texttt{ReverseArr}(ARR). Returns a reversed array \\
RemoveL & \texttt{RemoveL}(\underline{ARR}, NUM). Removes the last occurrence of given number* in array \\
RemoveF & \texttt{RemoveF}(\underline{ARR}, NUM). Removes the first occurrence of given number* in array \\
LastIndOf & {\em NUM} = \texttt{LastIndOf}(ARR, NUM). Returns the last index of given number* in array \\
FirstIndOf & {\em NUM} = \texttt{FirstIndOf}(ARR, NUM). Returns the first index of given number* in array \\
IsInArr & {\em BOOL} = \texttt{IsInArr}(ARR, NUM). Returns whether a given number is in array \\ \hline

{\bf GroupB}  &   \\ \hline       
AddArrays &  {\em ARR} = \texttt{AddArrays}(ARR, ARR). Adds corresponding elements of two arrays together \\
MultArrays & {\em ARR} = \texttt{MultArrays}(ARR, ARR). Multiplies corresponding elements of two arrays together \\
Sum       &  {\em NUM} = \texttt{Sum}(ARR). Returns sum of elements of an array            \\
SumOfSq &    {\em NUM} = \texttt{SumOfSq}(ARR). Returns sum of each element squared in array   \\
DotProd &    {\em NUM} = \texttt{DotProd}(ARR, ARR). Returns dot product of two arrays \\
MatVecMult & {\em ARR} = \texttt{MatVecMult}(AoA, ARR). Returns result of a matrix-vector multiply \\
AddToArr &   {\em ARR} = \texttt{AddToArr}(ARR, NUM). Adds a number to each element of an array \\
SubFromArr & {\em ARR} = \texttt{SubFromArr}(ARR, NUM). Subtracts a number from each element of an array  \\
ScaleArr &   {\em ARR} = \texttt{ScaleArr}(ARR, NUM). Multiplies each element of an array by a number \\
ScaledSum &  {\em NUM} = \texttt{ScaledSum}(ARR, NUM). Multiplies each element of an array by a number and sums the result \\ \hline

{\bf GroupC} & \\ \hline
CountEQ &   {\em NUM} = \texttt{CountEQ}(ARR). Returns number of elements equal to a given number in array \\
CountLT &   {\em NUM} = \texttt{CountLT}(ARR). Returns number of elements less than a given number in array\\   
CountGT &   {\em NUM} = \texttt{CountGT}(ARR). Returns number of elements greater than a given number in array\\   
SumIfLT &   {\em NUM} = \texttt{SumIfEQ}(ARR). Returns the sum of elements less than a given number in array \\      
SumIfGT &   {\em NUM} = \texttt{SumIfGT}(ARR). Returns the sum of elements greater than a given number in array \\         
SumIfEQ &   {\em NUM} = \texttt{SumIfEQ}(ARR). Returns the sum of elements equal to a given number in array \\      
ScaledAvg &    {\em NUM} = \texttt{ScaledAvg}(ARR). Returns the average of array elements scaled by a number \\
Sum      &     {\em NUM} = \texttt{Sum}(ARR). Returns sum of elements of an array \\
Len      &     {\em NUM} = \texttt{Len}(ARR). Returns the length of array \\
Avg      &     {\em NUM} = \texttt{Avg}(ARR). Returns the average of array \\ \hline
\end{tabular}

\caption{GroupA, GroupB, and GroupC problems. Symbol * indicates that number must be present in array}
\label{tab:fullproblems}
}
\vspace{-0.5in}
\end{table}

\vfill

\pagebreak[4]

\section*{Appendix B: Composition Graphs}
\label{aad:appendixB}

Table~\ref{tab:fullcallees} lists caller-callee relationships for all 3 groups of problems.

\begin{table}[h]
\footnotesize{
\begin{tabular}{l| c c c c c c c c c c} \hline
{\bf GroupA} \\ 

        & \rotatebox{90}{ Max}  & \rotatebox{90}{ Min}  & \rotatebox{90}{ SortDesc}     & \rotatebox{90}{ SortAsc}      & \rotatebox{90}{ ReverseArr} & \
\rotatebox{90}{ RemoveL}      & \rotatebox{90}{ RemoveF}      & \rotatebox{90}{ LastIndOf}    & \rotatebox{90}{ FirstIndOf}   & \rotatebox{90}{ IsInArr}\
    \\ \hline
 Max    &       &       &       &       &       &       &       &       &       &       \\
 Min    &       &       &       &       &       &       &       &       &       &       \\
 SortDesc       &57     &14     &       &43     &43     &14     &57     &       &       &       \\
 SortAsc        &4      &8      &94     &       &94     &6      &8      &       &       &18     \\
 ReverseArr     &       &       &       &       &       &       &       &       &       &       \\
 RemoveL        &2      &1      &       &       &26     &       &22     &79     &       &10     \\
 RemoveF        &1      &1      &       &       &59     &61     &       &       &43     &14     \\
 LastIndOf      &       &2      &       &       &       &       &       &       &       &       \\
 FirstIndOf     &       &       &       &       &       &       &2      &       &       &       \\
 IsInArr        &2      &3      &       &       &3      &1      &18     &61     &3      &       \\ \hline \hline

{\bf GroupB} \\ 

        & \rotatebox{90}{AddArrays}     & \rotatebox{90}{MultArrays}    & \rotatebox{90}{Sum}   & \rotatebox{90}{SumOfSq}       & \rotatebox{90}{DotProd}       & \rotatebox{90}{MatVecMult}    & \rotatebox{90}{AddToArr}      & \rotatebox{90}{SubFromArr}    & \rotatebox{90}{ScaleArr}      & \rotatebox{90}{ScaledSum}     \\ \hline
AddArrays       &       &       &2      &       &       &       &       &       &       &       \\
MultArrays      &       &       &       &       &       &       &       &       &       &       \\
Sum     &       &       &       &       &       &       &       &       &       &       \\
SumOfSq &3      &32     &29     &       &70     &       &3      &2      &2      &5      \\
DotProd &       &100    &93     &       &       &       &1      &1      &1      &8      \\
MatVecMult      &       &       &       &       &100    &       &13     &       &       &       \\
AddToArr        &46     &       &       &       &       &       &       &2      &       &       \\
SubFromArr      &2      &       &1      &1      &2      &       &86     &       &       &2      \\
ScaleArr        &       &49     &       &       &       &       &       &       &       &       \\
ScaledSum       &       &       &100    &       &       &       &       &       &1      &       \\ \hline \hline

{\bf GroupC} \\ 

        & \rotatebox{90}{CountEQ}       & \rotatebox{90}{CountLT}       & \rotatebox{90}{CountGT}       & \rotatebox{90}{SumIfLT}       & \rotatebox{90}{SumIfGT}       & \rotatebox{90}{SumIfEQ}       & \rotatebox{90}{ScaledAvg}     & \rotatebox{90}{Sum}   & \rotatebox{90}{Len}   & \rotatebox{90}{Avg}   \\ \hline
CountEQ &       &       &       &1      &       &1      &       &       &99     &       \\
CountLT &       &       &       &       &       &       &       &       &100    &       \\
CountGT &       &       &       &       &       &       &       &       &100    &       \\
SumIfLT &       &       &       &       &       &       &       &100    &       &       \\
SumIfGT &       &       &       &       &       &       &       &100    &       &       \\
SumIfEQ &       &       &       &       &       &       &       &100    &       &       \\
ScaledAvg       &       &       &       &       &       &       &       &12     &12     &88     \\
Sum     &       &       &       &       &       &       &       &       &       &       \\
Len     &       &       &       &       &       &       &       &4      &       &       \\
Avg     &       &       &       &       &       &       &37     &64     &64     &       \\ \hline 
\end{tabular}
\caption{Caller-callee relationships showing function composition for GroupA, GroupB, and GroupC problems.}
\label{tab:fullcallees}
}
\end{table}

\vfill

\pagebreak[4]

\section*{Appendix C: Parent-Child Graphs}
\label{aad:appendixC}

Table~\ref{tab:fullparents} lists parent-child relationships for all 3 groups of problems.

\begin{table}[h]
\footnotesize{
\begin{tabular}{l| c c c c c c c c c c} \hline
{\bf GroupA} \\ 

        & \rotatebox{90}{Max}   & \rotatebox{90}{Min}   & \rotatebox{90}{SortDesc}      & \rotatebox{90}{SortAsc}       & \rotatebox{90}{ReverseArr}    & \rotatebox{90}{RemoveL}       & \rotatebox{90}{RemoveF}       & \rotatebox{90}{LastIndOf}     & \rotatebox{90}{FirstIndOf}    & \rotatebox{90}{IsInArr}       \\ \hline
Max     &       &66     &       &       &       &       &       &       &       &       \\
Min     &68     &       &       &       &       &       &       &       &       &       \\
SortDesc        &       &       &       &       &       &       &       &       &       &       \\
SortAsc &       &       &       &       &       &       &       &       &       &       \\
ReverseArr      &       &       &       &       &       &       &       &       &       &       \\
RemoveL &       &       &       &       &       &       &       &       &       &       \\
RemoveF &       &       &       &       &       &       &       &       &       &       \\
LastIndOf       &       &       &       &       &       &       &       &       &77     &       \\
FirstIndOf      &       &       &       &       &       &       &       &99     &       &       \\
IsInArr &       &       &       &       &       &       &       &       &       &       \\ \hline \hline

{\bf GroupB} \\ 

        & \rotatebox{90}{AddArrays}     & \rotatebox{90}{MultArrays}    & \rotatebox{90}{Sum}   & \rotatebox{90}{SumOfSq}       & \rotatebox{90}{DotProd}       & \rotatebox{90}{MatVecMult}    & \rotatebox{90}{AddToArr}      & \rotatebox{90}{SubFromArr}    & \rotatebox{90}{ScaleArr}      & \rotatebox{90}{ScaledSum}     \\ \hline
AddArrays       &       &       &       &       &       &       &       &       &       &       \\
MultArrays      &       &       &       &       &       &       &       &       &       &       \\
Sum     &       &       &       &       &       &       &       &       &       &       \\
SumOfSq &       &       &       &       &       &       &       &       &       &       \\
DotProd &       &       &       &       &       &       &       &       &       &       \\
MatVecMult      &       &       &       &       &       &       &       &       &       &       \\
AddToArr        &       &       &       &       &       &       &       &       &       &       \\
SubFromArr      &       &       &       &       &       &       &       &       &       &       \\
ScaleArr        &       &       &       &       &       &       &15     &13     &       &       \\
ScaledSum       &       &       &       &       &       &       &       &       &       &       \\ \hline \hline

{\bf GroupC} \\ 
        & \rotatebox{90}{CountEQ}       & \rotatebox{90}{CountLT}       & \rotatebox{90}{CountGT}       & \rotatebox{90}{SumIfLT}       & \rotatebox{90}{SumIfGT}       & \rotatebox{90}{SumIfEQ}       & \rotatebox{90}{ScaledAvg}     & \rotatebox{90}{Sum}   & \rotatebox{90}{Len}   & \rotatebox{90}{Avg}   \\ \hline
CountEQ &       &38     &33     &1      &6      &19     &       &       &       &       \\
CountLT &43     &       &33     &2      &3      &16     &       &       &       &       \\
CountGT &43     &36     &       &1      &3      &15     &       &       &       &       \\
SumIfLT &2      &       &       &       &41     &56     &       &       &       &       \\
SumIfGT &1      &1      &       &43     &       &53     &1      &       &       &       \\
SumIfEQ &       &1      &       &42     &53     &       &1      &       &       &       \\
ScaledAvg       &       &       &       &22     &25     &34     &       &       &       &       \\
Sum     &       &       &       &       &       &       &       &       &       &       \\
Len     &       &       &       &       &       &       &       &5      &       &       \\
Avg     &       &       &       &       &       &       &       &       &       &       \\ \hline 
\end{tabular}
\caption{Parent-Child relationships for GroupA, GroupB, and GroupC problems.}
\label{tab:fullparents}
}
\end{table}
\vfill

\pagebreak[4]

\section*{Appendix D: Code Examples}
\label{aad:appendixD}

Code for each problem is given below. 
For brevity, only the main function is listed 
(i.e., the entire call tree is not listed for each solution).
It should be noted that there are often many other solutions 
found in addition to the one shown.

\subsection{GroupA Problems}

\codelistsize

\bigbreak
\begin{verbatim}
def Max(arg0):
        arr_10 = arg0.copy()
        num_11 = arr_10[0]
        for num_12 in tuple(arr_10):
                bool_14 = (num_12 > num_11)
                if (bool_14):
                        num_11 = num_12
                num_12 = num_11
        return (num_12)
\end{verbatim}

\bigbreak
\begin{verbatim}
def Min(arg0):
        arr_10 = arg0.copy()
        num_11 = arr_10[0]
        for num_12 in tuple(arr_10):
                bool_14 = (num_12 < num_11)
                if (bool_14):
                        num_11 = num_12
                num_12 = num_11
        return (num_12)
\end{verbatim}

\bigbreak
\begin{verbatim}
def SortDesc(arg0):
        arr_10 = arg0.copy()
        arr_15 = list()
        for num_13 in tuple(arr_10):
                num_16 = Max(arr_10)
                arr_15.append(num_16)
                arr_10 = RemoveF(arr_10, num_16)
        return (arr_15)
\end{verbatim}

\bigbreak
\begin{verbatim}
def SortAsc(arg0):
        arr_10 = arg0.copy()
        arr_19 = SortDesc(arr_10)
        arr_13 = ReverseArr(arr_19)
        arr_12 = arr_13.copy()
        return (arr_12)
\end{verbatim}

\bigbreak
\begin{verbatim}
def ReverseArr(arg0):
        arr_10 = arg0.copy()
        arr_17 = list()
        for num_12 in tuple(arr_10):
                num_13 = arr_10.pop()
                arr_17.append(num_13)
        return (arr_17)
\end{verbatim}

\bigbreak
\begin{verbatim}
def RemoveL(arg0, arg1):
        arr_10 = arg0.copy()
        num_11 = arg1
        num_15 = LastIndOf(arr_10, num_11)
        num_14 = arr_10.pop(num_15)
        bool_12 = (num_11 < num_14)
        return (arr_10)
\end{verbatim}

\bigbreak
\begin{verbatim}
def RemoveF(arg0, arg1):
        arr_10 = arg0.copy()
        num_11 = arg1
        num_17 = FirstIndOf(arr_10, num_11)
        num_16 = arr_10.pop(num_17)
        bool_12 = (num_17 < num_16)
        return (arr_10)
\end{verbatim}

\bigbreak
\begin{verbatim}
def LastIndOf(arg0, arg1):
        arr_10 = arg0.copy()
        num_11 = arg1
        for num_14, num_15 in enumerate(tuple(arr_10)):
                bool_17 = (num_15 == num_11)
                if (bool_17):
                        arr_10.append(num_14)
        num_12 = arr_10.pop()
        return (num_12)
\end{verbatim}

\bigbreak
\begin{verbatim}
def FirstIndOf(arg0, arg1):
        arr_10 = arg0.copy()
        num_11 = arg1
        arr_17 = list()
        for num_13, num_14 in enumerate(tuple(arr_10)):
                bool_16 = (num_14 == num_11)
                if (bool_16):
                        arr_17.append(num_13)
        num_12 = arr_17[0]
        return (num_12)
\end{verbatim}

\bigbreak
\begin{verbatim}
def IsInArr(arg0, arg1):
        arr_10 = arg0.copy()
        num_11 = arg1
        for num_12 in tuple(arr_10):
                bool_14 = (num_12 == num_11)
                if (bool_14):
                        num_11 = arr_10[-1]
                bool_13 = (num_12 == num_11)
        return (bool_13)
\end{verbatim}
\normalsize
\bigbreak

\subsection{GroupB Problems}

\codelistsize
\bigbreak
\begin{verbatim}
def AddArrays(arg0, arg1):
        arr_10 = arg0.copy()
        arr_11 = arg1.copy()
        for num_13 in tuple(arr_10):
                num_14 = arr_11.pop(0)
                num_15 = num_13 + num_14
                arr_11.append(num_15)
        return (arr_11)
\end{verbatim}

\bigbreak
\begin{verbatim}
def MultArrays(arg0, arg1):
        arr_10 = arg0.copy()
        arr_11 = arg1.copy()
        for num_13 in tuple(arr_10):
                num_14 = arr_11.pop(0)
                num_16 = arr_10.pop(0)
                num_15 = num_16 * num_14
                arr_11.append(num_15)
        return (arr_11)
\end{verbatim}

\bigbreak
\begin{verbatim}
def Sum(arg0):
        arr_10 = arg0.copy()
        num_14 = 0
        for num_12 in tuple(arr_10):
                num_14 = num_12 + num_14
        return (num_14)
\end{verbatim}

\bigbreak
\begin{verbatim}
def SumOfSq(arg0):
        arr_10 = arg0.copy()
        arr_14 = arr_10.copy()
        num_12 = DotProd(arr_14, arr_14)
        return (num_12)
\end{verbatim}

\bigbreak
\begin{verbatim}
def DotProd(arg0, arg1):
        arr_10 = arg0.copy()
        arr_11 = arg1.copy()
        arr_14 = MultArrays(arr_10, arr_11)
        num_13 = Sum(arr_14)
        return (num_13)
\end{verbatim}

\bigbreak
\begin{verbatim}
def MatVecMult(arg0, arg1):
        arr_of_arr10 = arg0
        arr_11 = arg1.copy()
        arr_16 = list()
        for arr_15 in tuple(arr_of_arr10):
                num_17 = DotProd(arr_11, arr_15)
                arr_16.append(num_17)
        arr_14 = arr_16.copy()
        return (arr_14)
\end{verbatim}

\bigbreak
\begin{verbatim}
def AddToArr(arg0, arg1):
        arr_10 = arg0.copy()
        num_11 = arg1
        for num_12 in tuple(arr_10):
                num_15 = arr_10.pop(0)
                num_14 = num_11 + num_15
                arr_10.append(num_14)
        return (arr_10)
\end{verbatim}

\bigbreak
\begin{verbatim}
def SubFromArr(arg0, arg1):
        arr_10 = arg0.copy()
        num_11 = arg1
        for num_12 in tuple(arr_10):
                num_13 = arr_10.pop(0)
                num_14 = num_13 - num_11
                arr_10.append(num_14)
        return (arr_10)
\end{verbatim}

\bigbreak
\begin{verbatim}
def ScaleArr(arg0, arg1):
        arr_10 = arg0.copy()
        num_11 = arg1
        arr_15 = list()
        for num_12 in tuple(arr_10):
                num_14 = num_12 * num_11
                arr_15.append(num_14)
        return (arr_15)
\end{verbatim}

\bigbreak
\begin{verbatim}
def ScaledSum(arg0, arg1):
        arr_10 = arg0.copy()
        num_11 = arg1
        num_13 = Sum(arr_10)
        num_12 = num_13 * num_11
        return (num_12)
\end{verbatim}
\normalsize
\bigbreak

\subsection{GroupC Problems}

\codelistsize
\bigbreak
\begin{verbatim}
def CountEQ(arg0, arg1):
        arr_10 = arg0.copy()
        num_11 = arg1
        arr_14 = list()
        for num_15 in tuple(arr_10):
                bool_17 = (num_11 == num_15)
                if (bool_17):
                        arr_14.append(num_15)
        arr_10.append(num_15)
        num_12 = Len(arr_14)
        return (num_12)
\end{verbatim}

\bigbreak
\begin{verbatim}
def CountLT(arg0, arg1):
        arr_10 = arg0.copy()
        num_11 = arg1
        arr_14 = list()
        for num_15 in tuple(arr_10):
                bool_17 = (num_11 > num_15)
                if (bool_17):
                        arr_14.append(num_15)
        arr_10.append(num_15)
        num_12 = Len(arr_14)
        return (num_12)
\end{verbatim}

\bigbreak
\begin{verbatim}
def CountGT(arg0, arg1):
        arr_10 = arg0.copy()
        num_11 = arg1
        arr_14 = list()
        for num_15 in tuple(arr_10):
                bool_17 = (num_15 > num_11)
                if (bool_17):
                        arr_14.append(num_15)
        arr_10.append(num_11)
        num_12 = Len(arr_14)
        return (num_12)
\end{verbatim}

\bigbreak
\begin{verbatim}
def SumIfLT(arg0, arg1):
        arr_10 = arg0.copy()
        num_11 = arg1
        arr_14 = list()
        for num_13 in tuple(arr_10):
                bool_15 = (num_11 <= num_13)
                if (bool_15):
                        num_13 = 0
                arr_14.append(num_13)
        num_12 = Sum(arr_14)
        return (num_12)
\end{verbatim}

\bigbreak
\begin{verbatim}
def SumIfGT(arg0, arg1):
        arr_10 = arg0.copy()
        num_11 = arg1
        arr_14 = list()
        for num_13 in tuple(arr_10):
                bool_15 = (num_11 >= num_13)
                if (bool_15):
                        num_13 = 0
                arr_14.append(num_13)
        num_12 = Sum(arr_14)
        return (num_12)
\end{verbatim}

\bigbreak
\begin{verbatim}
def SumIfEQ(arg0, arg1):
        arr_10 = arg0.copy()
        num_11 = arg1
        arr_14 = list()
        for num_13 in tuple(arr_10):
                bool_15 = (num_11 != num_13)
                if (bool_15):
                        num_13 = 0
                arr_14.append(num_13)
        num_12 = Sum(arr_14)
        return (num_12)
\end{verbatim}

\bigbreak
\begin{verbatim}
def ScaledAvg(arg0, arg1):
        arr_10 = arg0.copy()
        num_11 = arg1
        for num_13 in tuple(arr_10):
                num_14 = arr_10.pop(0)
                num_15 = num_14 * num_11
                arr_10.append(num_15)
        num_12 = Avg(arr_10)
        return (num_12)
\end{verbatim}

\bigbreak
\begin{verbatim}
def Sum(arg0):
        arr_10 = arg0.copy()
        num_17 = 0
        for num_12 in tuple(arr_10):
                num_17 = num_12 + num_17
        return (num_17)
\end{verbatim}

\bigbreak
\begin{verbatim}
def Len(arg0):
        arr_10 = arg0.copy()
        for num_14, num_15 in enumerate(tuple(arr_10)):
                pass
        num_13 = 1
        num_12 = num_13 + num_14
        return (num_12)
\end{verbatim}

\bigbreak
\begin{verbatim}
def Avg(arg0):
        arr_10 = arg0.copy()
        arr_20 = arr_10.copy()
        num_18 = Sum(arr_20)
        num_17 = Len(arr_10)
        num_13 = num_18 // num_17
        num_12 = num_13
        return (num_12)
\end{verbatim}

\normalsize

\pagebreak[4]

\section*{Appendix E: Source Code \& Result Files}
\label{aad:appendixE}

\setcounter{subsection}{0}
\renewcommand*{\theHsubsection}{chY.\the\value{subsection}}
\subsection{Source Code}
Please contact authors for source code (main.py).
It can be run with Python
3.6 or later using the following command:

\bigbreak
\texttt{python main.py groupID }
\bigbreak

where, \texttt{groupID} is 1, 2, or 3,
for GroupA, GroupB, and GroupC, respectively.
The run prints progress on stdout and generates a report at the end (containing least-complex results, 
callees, parents, a stat record for each solution, etc.) 
in the current directory,
a checkpoint in the \texttt{./chkpts} directory,
and a detailed log file for each rank in the \texttt{./log.\emph{nodename}} directory.
On machines with fewer cores (than 112 we used), 
number of epochs \emph{must} be increased proportionately for all solutions to be found.

After one or more such runs, one or more checkpoints can be 
read and least-complex solutions can be constructed for reporting purposes
by running

\bigbreak
\texttt{python main.py groupID 1}
\bigbreak

Notice that least complex code produced in this step is composed
from the least complex result found for each problem and the 
composed code is not currently tested using a Checker.
This composed code is for reporting purposes only and 
must be inspected by the user. 

\subsection{Result Files}
Please contact authors for result files.

\end{document}